\documentclass{llncs}
\usepackage{amsmath,amssymb}
\usepackage{graphicx,array}
\usepackage[all]{xy}
\renewcommand{\aa}{\boldsymbol{a}}
\newcommand{\bb}{\boldsymbol{b}}
\newcommand{\cc}{\boldsymbol{c}}

\newcommand{\hh}{\boldsymbol{h}}
\newcommand{\mm}{\boldsymbol{m}}
\newcommand{\uu}{\boldsymbol{u}}
\newcommand{\xx}{\boldsymbol{x}}
\newcommand{\yy}{\boldsymbol{y}}
\newcommand{\CC}{\boldsymbol{C}}

\newcommand{\XX}{\boldsymbol{X}}
\newcommand{\YY}{\boldsymbol{Y}}
\newcommand{\ZZ}{\boldsymbol{Z}}
\newcommand{\mmu}{\boldsymbol{\mu}}
\newcommand{\GGamma}{\boldsymbol{\Gamma}}
\newcommand{\SSigma}{\boldsymbol{\Sigma}}
\newcommand{\Acal}{\mathcal{A}}

\newcommand{\Scal}{\mathcal{S}}

\newcommand{\cov}{\mathrm{cov}}
\newcommand{\drm}{\mathrm{d}}
\newcommand \EI {\mathrm{EI}}
\newcommand{\esp}{\mathbb{E}}
\newcommand{\GP}{\mathcal{GP}}

\newcommand \N {\mathcal N}
\newcommand \qEI {q\textnormal{-}\mathrm{EI}}
\newcommand \trans {^\top}
\newcommand \tr {\mathrm{tr}}
\newcommand{\comp}[2]{#1\circ #2}
\newcommand{\couple}[2]{\left(#1, #2\right)}

\newcommand{\Rset}{\bbbr}
\newcommand{\diff}[2]{d_{#1}\left[#2\right]}
\newcommand{\gan}{g_1}
\newcommand{\gbn}{g_2}
\newcommand{\gcn}{g_3}
\newcommand{\gdn}{g_4}
\newcommand{\gen}{g_5}
\newcommand{\gfn}{g_{6}}
\newcommand{\ggn}{g_7}
\newcommand{\ghn}{g_8}

\newcommand{\gcouple}{G}
\begin{document}

\title{Differentiating the 
multipoint Expected Improvement for 
optimal batch design}

\author{S\'ebastien Marmin \inst{1} \inst{2} \inst{3}%
\and Cl\'ement Chevalier  \inst{4}\inst{5} \and David Ginsbourger
\inst{6}\inst{1}}

\institute{
IMSV, Department of Mathematics and Statistics, University of Bern, Switzerland\\
\and
Institut de Radioprotection et de S\^uret\'e Nucl\'eaire, Cadarache, France\\
\and 
\'Ecole Centrale de Marseille, France\\ 
\and
Institute of Statistics, University of Neuch\^atel, Switzerland\\
\and 
Institute of Mathematics, University of Zurich, Switzerland\\
\and 
Idiap Research Institute, Martigny, Switzerland
}

\maketitle
%
\makeatletter

\makeatother
\begin{abstract}
This work deals with parallel optimization of 
expensive objective functions which are modelled as 
sample realizations of Gaussian processes. 
The study is formalized as a Bayesian optimization 
problem, or continuous multi-armed bandit problem, 
where a batch of $q > 0$ arms is pulled in parallel 
at each iteration. 
Several algorithms have been developed for choosing 
batches by trading off exploitation and exploration. 
As of today, the maximum Expected Improvement (EI) and 
Upper Confidence Bound (UCB) selection rules appear 
as the most prominent approaches for batch selection. 
Here, we build upon recent work on the multipoint 
Expected Improvement criterion, for which an 
analytic expansion relying on Tallis' formula was 
recently established. 
The computational burden of this selection rule 
being still an issue in application, we derive a 
closed-form expression for the gradient of the 
multipoint Expected Improvement, which aims at 
facilitating its maximization using gradient-based 
ascent algorithms. 
Substantial computational savings are shown in 
application. In addition, our algorithms are tested  
numerically and compared to state-of-the-art 
UCB-based batch-sequential algorithms. 
Combining starting designs relying on UCB with 
gradient-based EI local optimization finally 
appears as a sound option for batch design in 
distributed Gaussian Process optimization. 
\end{abstract}
\section{Introduction}

Global optimization of deterministic functions 
under a drastically limited evaluation budget 
is a topic of growing interest with important 
industrial applications. 
Dealing with such expensive black-box 
simulators is typically addressed through the 
introduction of surrogate models that are  
used both for reconstructing the objective 
function  and guiding parsimonious evaluation 
strategies.
This approach is used in various scientific 
communities and referred to as Bayesian optimization, 
but also as kriging-based or multi-armed bandit optimization  
\cite{Rasmussen,Brochu:2010c,srinivas2012information}
\cite{Jones,ginsbourger:2010:kws,villemonteix:2009:iago,frazier:2012}.
Among such Gaussian process optimization methods, two 
concepts of algorithm relying on sequential maximization 
of infill sampling criteria are particularly popular in 
the literature. 
In the EGO algorithm of \cite{Jones}, the sequence of decisions (of where to evaluate the objective function at each iteration) is guided by the Expected Improvement (EI) criterion \cite{Mockus}, which is known to be one-step lookahead optimal \cite{ginsb_horizon}. 
On the other hand, the Upper Confidence Bound (UCB) algorithm \cite{auer2002finite} maximizes sequentially a well-chosen 
kriging quantile, that is, a quantile of the pointwise posterior Gaussian process distribution. 
Similarly to EI \cite{vazquez2010convergence,Bull2011}, the consistency of the algorithm has been established and rates of convergence have been obtained \cite{srinivas2012information}. 

Recently, different methods inspired from the two latter algorithms have been proposed to deal with the typical case where $q>1$ CPUs 
are available. Such synchronous distributed methods provide at each 
iteration a batch of $q$ points which can be evaluated in parallel. 
For instance, \cite{desautels2012parallelizing} generalizes the UCB algorithm to a batch-sequential version by maximizing 
kriging quantiles and assuming dummy responses equal to 
the posterior mean of the Gaussian process.
This approach can be compared with the so-called Kriging Believer strategy of \cite{ginsbourger:2010:kws} where each batch is obtained 
by sequentially maximizing the one-point EI under the assumption that the previously chosen points have a response equal to their Kriging mean. 
Originally, the strategies suggested in \cite{ginsbourger:2010:kws} were introduced to cope with the difficulty to evaluate and maximize the multipoint Expected Improvement ($\qEI$) \cite{schonlau:phd}, 
which is the generalization of EI known to be one-batch lookahead optimal \cite{chevalier:phd,ginsb_horizon}. 
One of the bottlenecks for $\qEI$ maximization was that it was until recently evaluated through Monte-Carlo simulations \cite{ginsbourger:2010:kws}, a reason that motivated \cite{frazier:2012} to propose a stochastic gradient algorithm for 
its maximization. 
Now, \cite{chevalier:multipointEI} established a closed-form expression enabling to compute $\qEI$ at any batch of $q$ points without appealing to Monte-Carlo simulations. 
However, the computational complexity involved 
to compute the criterion is still high and quickly grows with 
$q$. Besides, little has been published about the 
difficult maximization of the $\qEI$ itself, which is an optimization problem in dimension $q d$, where $d$ is the number of 
input variables. 

In this work, we contribute to the latter problem 
by giving an analytical gradient of $\qEI$, in the space of dimension $q d$. Such a gradient is meant to simplify the local maximization of $\qEI$ using gradient-based ascent algorithms. 
Closed-form expressions of $\qEI$ and its gradient have been implemented in the DiceOptim R package 
\cite{ginsbourger:dicekrigingoptim}, together with a 
multistart BFGS algorithm for maximizing $\qEI$. 
In addition, we suggest to use results of the BUCB algorithm as initial batches in multistart gradient-based ascents. 
These starting batches are shown to yield good 
local optima for $\qEI$.  
This article is organized 
as follows. Section~\ref{sec:2} quickly recalls 
the basics of Gaussian process modeling and 
the closed-form expression of $\qEI$ 
obtained in \cite{chevalier:multipointEI}.  
Section~\ref{sec:3} details the analytical $\qEI$ gradient. 
Finally, numerical experiments comparing the performances 
of the $\qEI$ maximization-based strategy and 
the BUCB algorithms are provided and discussed 
in Section~\ref{sec:4}. For readability and 
conciseness, the most technical details about $\qEI$ gradient calculation are sent in Appendix. 

\section{General Context}
\label{sec:2}
Let $f:\xx\in D\subset\Rset^d\longrightarrow \Rset$ 
be a real-valued  function defined on a compact subset $D$ 
of $\Rset^d, d\geq 1$. Throughout this 
article, we assume that we dispose of a set of 
$n$ evaluations of $f$, 
$\Acal_n=\left(\xx_{1:n} := 
\{\xx_{1},\ldots,\xx_{n}\},
\yy_{1:n}=(f(\xx_{1}),\ldots,f(\xx_{n}))\trans\right)$, and that  
our goal is to evaluate $f$ at well-chosen batches of $q$ 
points in order to globally maximize it.  
Following each batch of evaluations, 
we observe $q$ deterministic scalar responses, or rewards, 
$y_{n+1}=f(\xx_{n+1}),\ldots,y_{n+q}=f(\xx_{n+q})$. 
We use past observations in order to 
carefully choose the next $q$ observation locations, aiming in the end to 
minimize the one-step lookahead regret 
$f(\xx^*)-t_{n+q}$, where $\xx^*$ is a maximizer of $f$ and 
$t_i = \max_{j=1,\ldots,i}(f(\xx_{j}))$. 
In this section, we first define the Gaussian 
process (GP) surrogate model used to make the decisions. Then we introduce the $\qEI$ which is 
the optimal one-batch lookahead criterion (see, e.g.,  \cite{ginsb_horizon,bect:2011:stco,frazier:2008:kgp} for a definition and \cite{ginsb_horizon,chevalier:phd} for a proof). 

\subsection{Gaussian process modeling}

The objective function $f$ is a priori assumed 
to be a sample from a Gaussian 
process $Y\sim\GP(\mu,C)$, 
where $\mu(\cdot)$ and $C(\cdot,\cdot)$ are 
respectively the mean and covariance function 
of $Y$. 
At fixed $\mu(\cdot)$ and $C(\cdot,\cdot)$, 
conditioning $Y$ on 
the set of observations $\Acal_n$ yields 
a GP posterior $Y(\xx)|\Acal_n \sim \GP(\mu_n,C_n)$  with:
\begin{align}
\label{krig:mean}
\mu_n(\xx) &= \mu(\xx) + \cc_n(\xx)\trans 
\CC_n^{-1}(\yy_{1:n} - \mu(\xx_{1:n})),\text{ and}\\
\label{krig:cov}
C_n(\xx,\xx^{'}) &= C(\xx,\xx^{'}) - \cc_n(\xx)\trans\CC_n^{-1}\cc_n(\xx^{'}),
\end{align}
where $\cc_n(\xx) = 
(C(\xx,\xx_{i}))_{1\le i\le n}$, and $\CC_n = (C(\xx_{i},\xx_{j}))_{1\le i,j\le n}$.
Note that, in realistic application settings, the mean and the covariance  $\mu$ and $C$ of the prior are assumed to depend on several parameters which require to be estimated. 
The results presented in this article and their implementations 
in the \verb|R| package \verb|DiceOptim| are compatible with this more general case. 
More detail about Equations~\ref{krig:mean}, \ref{krig:cov} 
with or without trend and covariance parameter estimation can be found in \cite{ginsbourger:dicekrigingoptim} and is omitted here for conciseness.

\subsection{The Multipoint Expected Improvement criterion}

The Multipoint Expected Improvement ($\qEI$) selection rule consists in maximizing, over all possible batches of 
$q$ points, the following criterion, which depends on 
a batch $\XX=(\xx_{n+1},\ldots,\xx_{n+q})\in D^q$:
\begin{equation}
\EI(\XX) = \esp \left[\left(\max Y(\XX) - T_{n}\right)_+\right|\Acal_n],
\end{equation}
where $(\cdot)_+=\max(\cdot,0)$, and the threshold $T_{n}$ is the currently observed maximum of $Y$, i.e. 
$T_{n}=\max_{1\leq j\leq n}Y(\xx_{j})$. 
Recalling that 
 $Y(\XX)|\Acal_n \sim\N(\mmu_n(\XX),C_n(\XX,\XX))$,  
and denoting $Y(\XX) = (Y_1,\hdots,Y_q)\trans$, 
an analytic expression of $\qEI$ at 
locations $\XX$ over any threshold $T\in\Rset$ can be found in \cite{chevalier:multipointEI} and is 
reproduced here :
\begin{align}
\label{eq:qEIanalytic}
\EI(\XX)=&\sum_{k=1}^q\left((m_k-T)\Phi_{q,\Sigma^{(k)}}\left(-\mm^{(k)}\right)+\sum_{i=1}^q \Sigma_{ik}^{(k)} \varphi_{\Sigma_{ii}}\left(m_i^{(k)}\right)\Phi_{q-1,\Sigma^{(k)}_{|i}}\left(-\mm^{(k)}_{|i}\right)\right)
\end{align}
where 
$\varphi_{\sigma^2}(\cdot)$ and $\Phi_{p,\Gamma}(\cdot)$ are respectively the density function of the centered normal distribution with variance $\sigma^2$ and the $p$-variate cumulative distribution function (CDF) of the centered normal distribution with covariance $\Gamma$ ; 
$\mm = \esp(Y(\XX) | \Acal_n)$ and 
 $\Sigma=\cov(Y(\XX) | \Acal_n)$ are the 
 conditional mean vector and  covariance matrix of $Y(\XX)$ ; 
$\mm^{(k)}$ and $\Sigma^{(k)}$, 
$1\leq k \leq q$, are the conditional mean vector 
and covariance matrix of the affine 
transformation of $Y(\XX)$, $\ZZ^{(k)}=L^{(k)}Y(\XX)+\bb^{(k)}$, defined as 
$Z^{(k)}_j:=Y_j$ for $j\neq k$ and 
$Z_k^{(k)}:=T - Y_k$ ; 
and finally, 
for $(k,i) \in \{1,\ldots,q\}^2$, 
$\mm^{(k)}_{|i}$ and $\Sigma^{(k)}_{|i}$ are the mean vector and covariance matrix of the Gaussian vector $(\ZZ^{(k)}_{-i}|Z^{(k)}_i=0)$, the index $-i$ 
meaning that the $i^\text{th}$ component is removed.
\section{Gradient of the multipoint Expected Improvement}
\label{sec:3}

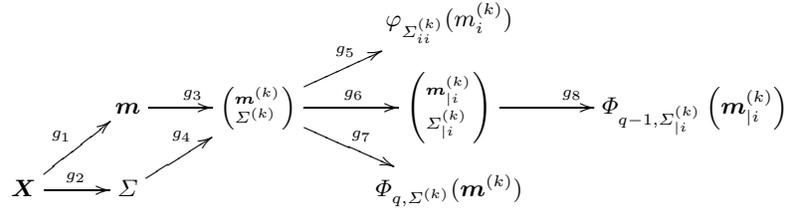
\begin{figure}
\begin{displaymath}
    \xymatrix@R=0.2cm{
              &            &                                        &\varphi_{\Sigma_{ii}^{(k)}}(m_i^{(k)})&\\
              &\mm \ar[r]^\gcn   &\left({{}^{\mm^{(k)}}_{\Sigma^{(k)}}}\right)  \ar[r]^\gfn\ar[ru]^\gen\ar[rd]^\ggn&\left({{}^{\mm_{|i}^{(k)}}_{\Sigma_{|i}^{(k)}}}\right) \ar[r]^\ghn&\Phi_{q-1,\Sigma_{|i}^{(k)}}\left(\mm_{|i}^{(k)}\right)\\
\XX\ar[ru]^\gan\ar[r]^\gbn&\Sigma\ar[ru]^\gdn& &\Phi_{q,\Sigma^{(k)}}(\mm^{(k)})&
  }
\end{displaymath}
\caption{Link between the 
different terms of Eq.~\eqref{eq:qEIanalytic} 
and the batch of points $\XX$}
\label{fig:Xdependence}
\end{figure}

In this section, we provide an
analytical formula for the gradient of $\qEI$. 
Getting such formula requires to carefully analyze the dependence of $\qEI$ written in Eq.~\eqref{eq:qEIanalytic} on the batch locations $\XX\in\Rset^{q\times d}$.
This dependence is summarized in Fig.~\ref{fig:Xdependence} and exhibits 
many chaining relations.
In the forthcoming multivariate calculations, 
we use the following notations.
Given two Banach spaces $E$ and $F$, and a differentiable function $g:E\rightarrow F$, the differential of $g$ at point $x$, written $\diff{x}{g}:E\rightarrow F$, is the bounded linear map that best approximate $g$ in the neighborhood of $x$. In the case where $E=\Rset^p$ and $F=\Rset$, it is well known that
$ \forall h\in E,  
\diff{x}{g}(h)= \langle \nabla g(x),h \rangle
$. More generally the differential can be written in terms of Jacobian matrices, matrix derivatives and/or matrix scalar products where $E$ and/or $F$ are $\Rset^p$ or $\Rset^{p\times p}$.
To simplify notations and handle the different indices in Eq.~\eqref{eq:qEIanalytic}, 
we fix the indices $i$ and $k$ and focus on differentiating the function $\EI^{(k)(i)}$, standing for
the generic term of the double sums in Eq.~\eqref{eq:qEIanalytic}. 
We can perform the 
calculation of 
$\diff{\XX}{\EI^{(k)(i)}}$
 by 
noticing that
$\EI^{(k)(i)}$ can be rewritten using the functions $g_j, 1\leq j\leq 8$ defined 
on Fig.~\ref{fig:Xdependence} as follows:
\begin{equation}
\label{eq:qEIki}
\EI^{(k)(i)}=\left(m_k-T\right) \cdot \comp{\ggn}{\gcouple} +  \comp{\gdn}{\gbn} \cdot \comp{\gen}{\gcouple} \cdot \comp{\ghn}{\comp{\gfn}{\gcouple}},
\end{equation}
where $\gcouple=\couple{\comp{\gcn}{\gan}}{\comp{\gdn}{\gbn}}$, $\circ$ is the 
composition operator and $\cdot$ the multiplication operator.
The differentiation then consists in applying classical differentiation formulas for 
products and compositions to Eq.~\eqref{eq:qEIki}. 
Proposition~\ref{propo:grad} summarizes the results. 
For conciseness, the formulae of the differentials involved in 
Eq.~\eqref{eq:grad} are justified in 
the Appendix. 
The calculations notably rely on the differential of a normal cumulative distribution function with respect to its covariance matrix obtained via Plackett's formula \cite{berman1987extension}.
{
\proposition \label{propo:grad} 
The differential of the multipoint Expected Inmprovement criterion of Eq.~\eqref{eq:qEIanalytic} is given by 
$\diff{\XX}{\EI}=\sum_{k=1}^q\sum_{i=1}^q \diff{\XX}{\EI^{(k)(i)}}$, 
with
\begin{align}
\label{eq:grad}
\diff{\XX}{\EI^{(k)(i)}} &=  \diff{\XX}{m_k}~.~\comp{\ggn}{\gcouple} + \left(m_k-T\right)~.~\comp{\diff{\gcouple(\XX)}{\ggn}}{\diff{\XX}{\gcouple}}  \\&+\comp{\diff{\gbn(\XX)}{\gdn}}{ \diff{\XX}{\gbn}}~.~\comp{\gen}{\gcouple} ~.~ \comp{\ghn}{\comp{\gfn}{\gcouple}}\nonumber\\
&+\comp{\gdn}{\gbn}~.~\comp{\diff{\gcouple(\XX)}{\gen}}{\diff{\XX}{\gcouple}} ~.~ \comp{\ghn}{\comp{\gfn}{\gcouple}}\nonumber\\
&+\comp{\gdn}{\gbn}~.~\comp{\gen}{\gcouple} ~.~ \comp{\comp{\diff{\gfn(\gcouple(\XX))}{\ghn}}{\diff{\gcouple(\XX)}{\gfn}}}{\diff{\XX}{\gcouple}}\nonumber,
\end{align}
where the $g_j$'s are the functions introduced  
in Fig.~\ref{fig:Xdependence}. The $g_j$'s and their respective differentials  
are as follow :
\smallskip
\begin{itemize}
\item $\gan$ : $\XX\in D^q\rightarrow \gan(\XX)=({\mu_n(\xx_j)})_{1\le j\le q}\in\Rset^q$, \\
$\diff{\XX}{\gan}(H) = (\langle \nabla\mmu_n({\xx_j}),H_{j,1:d}\trans \rangle)_{1\le j\le q}$,   \\
with $\nabla\mmu_n({\xx_j})=\nabla\mmu(\xx_j) + \left(\frac{\partial \cc_n(\xx_j)\trans}{\partial x_\ell}\right)_{\substack{1\le \ell\le d}}\CC_n^{-1}\left(\yy_{1:n}-\mu(\xx_{1:n})\right)$.
\smallskip
\item $\gbn$ : $\XX\in D^q\rightarrow \gbn(\XX)=({C_n(\xx_j,\xx_\ell)})_
{\substack{1\le j,\ell \le q}}\in\Scal_{++}^q$. $\Scal_{++}^q$ is the set of $q\times q$ positive definite matrices. \\
$\diff{\XX}{\gbn}(H) =  \left(\left\langle\nabla_{\xx} C_n(\xx_j,\xx_\ell),H_{j,1:d}\trans\right\rangle 
+ \left\langle\nabla_{\xx} C_n(\xx_\ell,\xx_j),H_{\ell,1:d}\trans\right\rangle\right)_{1\le j,\ell\le q}$, \\
with $\nabla_{\xx} C_n({\xx},\xx')=\nabla_{\xx} C(\xx,\xx') -\left(\frac{\partial \cc_n(\xx)\trans}{\partial x_p}\right)_{\substack{1\le p\le d}}\CC^{-1}_n\cc_n(\xx')$.

\smallskip
\smallskip
\item $\gcouple$ : $\XX\rightarrow \left(\mm^{(k)},\Sigma^{(k)}\right)$, 
$\diff{\XX}{\gcouple} = \left(L^{(k)}\diff{\XX}{\gan},L^{(k)}\diff{\XX}{\gbn}L^{(k)\top}\right)$.
\smallskip
\smallskip
\item $\ggn$ : $(\aa,\Gamma) \in \Rset^q\times\Scal_{++}^q\rightarrow \Phi_{q,\Gamma}(\aa)\in\Rset$, \\
$\diff{G(\XX)}{\ggn}(\hh,H) = \langle \hh,\nabla_{\xx}\Phi_{q,\Sigma^{(k)}}(\mm^{(k)})\rangle+\tr(H\nabla_\Sigma\Phi_{q,\Sigma^{(k)}}(\mm^{(k)}))$.    $\nabla_{\xx}\Phi_{q,\Sigma^{(k)}}$ and 
$\nabla_\Sigma\Phi_{q,\Sigma^{(k)}}$ are the gradient of the multivariate Gaussian CDF with respect to $\xx$ and to the covariance matrix, given in appendix.
\smallskip
\smallskip
\item $\gdn$ : $\Sigma\rightarrow \Sigma^{(k)}$, 
$\diff{\gbn(\XX)}{\gdn}(H) = L^{(k)} H L^{(k)\top}$.
\smallskip
\smallskip
\item $\gen$ : $(\aa,\Gamma) \in \Rset^q\times\Scal_{++}^q\rightarrow \varphi_{\Gamma_{ii}}(a_i)\in\Rset$, \\
$\diff{G(\XX)}{\gen}(\hh,H) = \left(-\frac{a_i}{\Gamma_{ii}}h_i + \frac{1}{2}\left(\frac{a_i^2}{\Gamma_{ii}^2}-\frac{1}{\Gamma_{ii}}\right)H_{ii}\right)\varphi_{\Gamma_{ii}}(a_i)$
\smallskip
\smallskip
\item $\gfn$ : $(\mm^{(k)},\Sigma^{(k)})\in \Rset^q\times\Scal_{++}^q \rightarrow (\mm^{(k)}_{|i},\Sigma^{(k)}_{|i})$, 

\begin{align}
\diff{\left(\mm^{(k)},\Sigma^{(k)}\right)}{\gfn}(h,H) = \left(\hh_{-i} -\frac{\hh_i}{\Sigma_{ii}^{(k)}}\SSigma_{-i,i}^{(k)} +\frac{m^{(k)}_i H_{ii}}{\Sigma_{ii}^{(k)2}}\SSigma_{-i,i}^{(k)}-\frac{m^{(k)}_i}{\Sigma_{ii}^{(k)}}H_{-i,i}~,\right.\nonumber\\\left. H_{-i,-i}+\frac{H_{ii}}{\Sigma_{ii}^{(k)2}}\SSigma_{-i,i}^{(k)}\SSigma_{-i,i}^{(k)\top}-\frac{1}{\Sigma_{ii}^{(k)}}H_{-i,i}\SSigma_{-i,i}^{(k)\top}-\frac{1}{\Sigma_{ii}^{(k)}}\SSigma_{-i,i}^{(k)}H_{-i,i}\trans\right)\nonumber
\end{align}
\smallskip
\smallskip
\item $\ghn$ : $(\aa,\Gamma) \in \Rset^{q-1}\times\Scal_{++}^{q-1}\rightarrow \Phi_{q-1,\Gamma}(\aa)\in\Rset$, \\
$\diff{\gfn(\gcouple(\XX))}{\ghn} = \langle \hh,\nabla_{\xx}\Phi_{q,\Sigma^{(k)}}(\mm^{(k)})\rangle+\tr(H\nabla_\Sigma\Phi_{q,\Sigma^{(k)}}(\mm^{(k)}))$.
\end{itemize}
}
The gradient of $\qEI$, relying on Eq.~\eqref{eq:grad} is implemented in the version $1.5$ of the DiceOptim R package \cite{D.Ginsbourger.etal2015}, together with a gradient-based local optimization algorithm. 
In the next section, we show that the analytical 
computation of the gradient offers substantial 
computational savings compared to 
numerical computation based on a finite-difference 
scheme. 
In addition, we investigate the performances of the batch-sequential EGO algorithm consisting in sequentially maximizing $\qEI$, and we compare it with the BUCB algorithm of 
\cite{desautels2012parallelizing}.
\section{Numerical tests}
\label{sec:4}

\subsection{Computation time}
\begin{figure}
   \begin{minipage}[t]{0.5\textwidth}
   \centering
      \includegraphics[trim= 4 18 35 59,width=\textwidth,clip=TRUE]{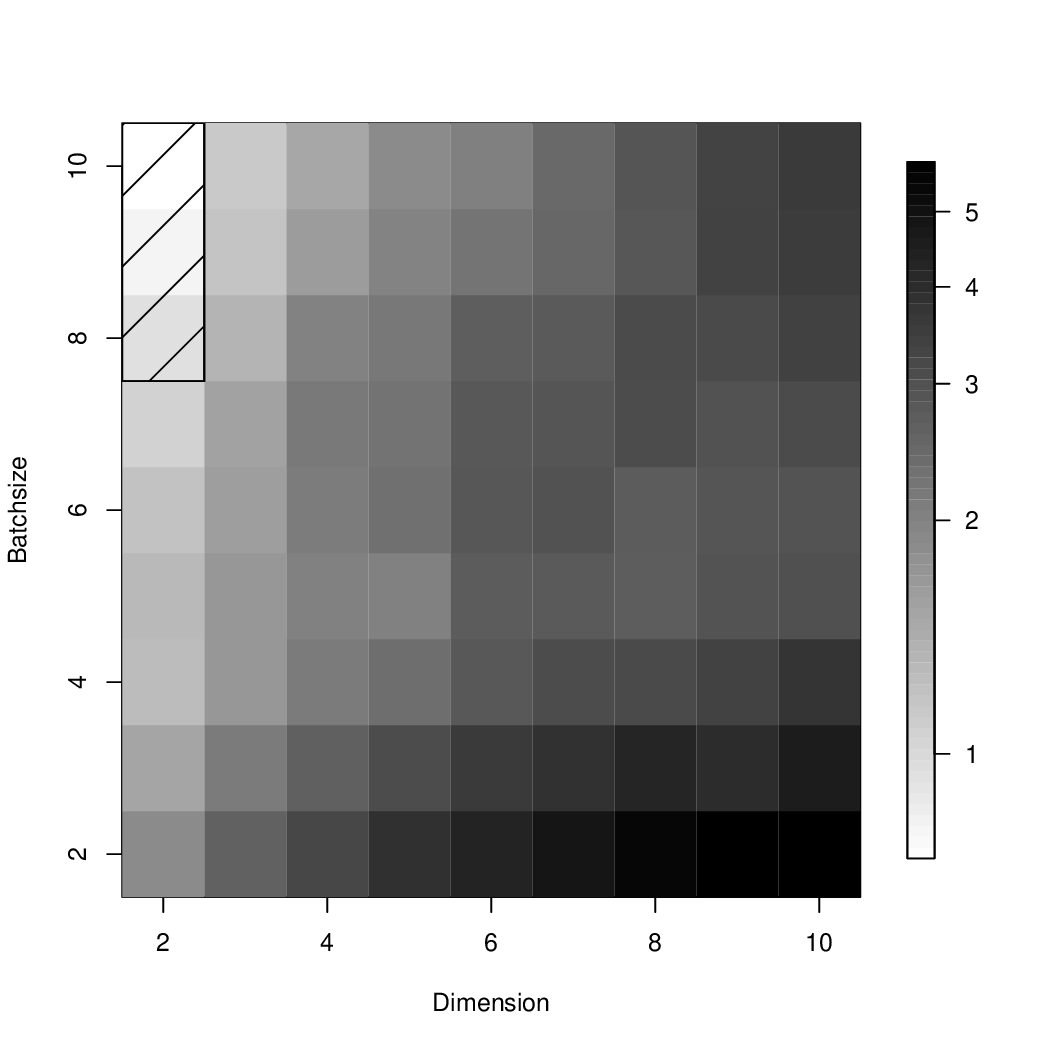}
      \caption{Ratio between computation times of the numerical and 
analytical gradient of $\qEI$ as a function of the dimension $d$ and the batch size $q$. 
The hatched area indicates 
a ratio below $1$.
}
\label{fig:computationtime}
   \end{minipage}
   \hspace{4mm}
   \begin{minipage}[t]{0.5\textwidth}
      \centering
      \includegraphics[trim= 4 18 30 59,width=\textwidth,clip=TRUE]{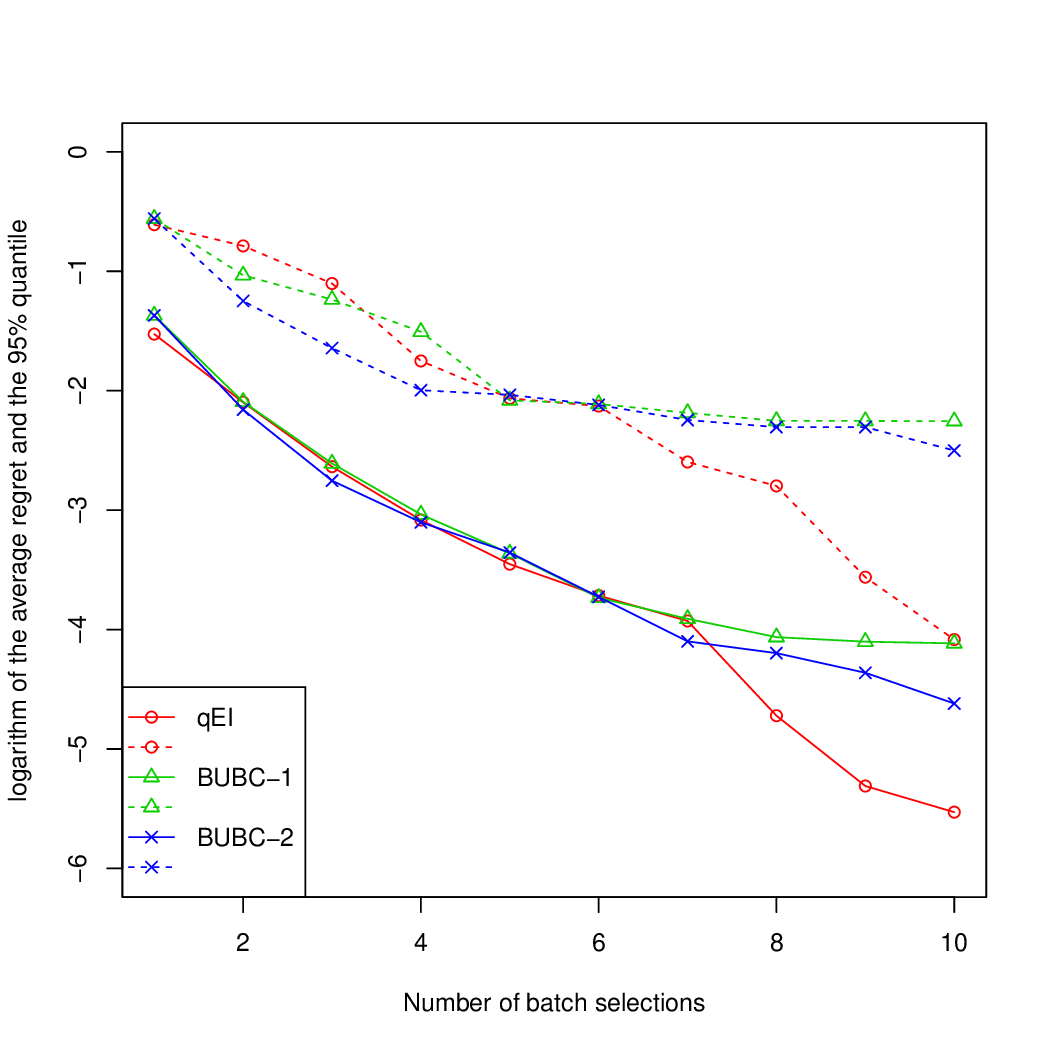}
      \caption{Logarithm of the average (plain lines) and 95\% quantile (dotted lines) of the 
      regret for three different batch-sequential optimization strategies (see Section~4.2 for detail).}
\label{fig:regret}
   \end{minipage}
\end{figure}

In this section, we illustrate the benefits -- in terms of computation time -- of using the analytical gradient formula 
of Section~\ref{sec:3}.
We compare computation times of gradients computed 
analytically and numerically, through finite differences 
schemes.
It is important to note that the computation 
of both $\qEI$ and its gradient 
(see, Eqs.~\eqref{eq:qEIanalytic},\eqref{eq:grad}) 
involve several calls to the cumulative distribution 
functions (CDF) of the multivariate normal distribution.  
The latter CDF is computed numerically with the  
algorithms of \cite{genz1992} wrapped in the mnormt R package 
\cite{mnormt}. 
In our implementation, computing this CDF turns out to be the main bottleneck in terms of computation time. 
The total number of calls to this CDF (be it in dimension $q$, $q-1$, $q-2$ or $q-3$) is summarized in Table~\ref{table:tps}.
\begin{table}
\centering
\begin{tabular}{r|c|c|c|c|c}
  &  $\Phi_{q-3}$ &   $\Phi_{q-2}$ & $\Phi_{q-1}$ & $\Phi_{q}$ & Total \ \\ 
  \hline
analytic $\qEI$ &  0 & 0 & $q^2$ & $q$& $O(q^2)$ \\ 
finite differences gradient & 0 &0 & $q(d+1) ~q^2$ & $q(d+1) ~q$& $O(dq^3)$\\
analytic gradient &  $q^2\frac{q(q-1)}{2}$  & $q\frac{q(q-1)}{2}+q^3$ & $q^2+2q^2$&$q$& $O(q^4)$
\end{tabular}
\vspace{0.3cm}
\caption{Total number of calls to the CDF of the multivariate Gaussian distribution for computing $\qEI$ or its gradient for a function with $d$ input variables.  
{ The last column gives the overall computational complexity.}}
\label{table:tps}
\end{table}
From this table, let us remark that the number of CDF calls does not depend on $d$ for the 
analytical $\qEI$ gradient and is proportional to $d$ for the numerical gradient. 
The use of the analytical gradient is thus expected to bring savings when $q$ is not too 
large compared to $d$. 
Figure~\ref{fig:computationtime} depicts the ratio of computation times between 
numerical and analytical gradient, as a function of $q$ and $d$. 
These were obtained by averaging the evaluation times of $\qEI$'s gradient at $10$ randomly-generated batches of size $q$ for a given function in dimension $d$ being a 
sample path of a GP with separable Mat\'ern(3/2) covariance function 
\cite{ginsbourger:dicekrigingoptim}. 
In the next section, we use the values $q = 6$ and $d = 5$ and we rely exclusively on the analytical $\qEI$ formula { which is now known to be faster}. 
\subsection{Tests}
\subsubsection{Experimental setup}

We now compare the performances 
of two parallel Bayesian optimization algorithm 
based, respectively, on the UCB approach of \cite{srinivas2012information} and on sequential $\qEI$ 
maximizations. 
We consider a minimization problem in dimension 
$d = 5$ where $n = 50$ evaluations are performed 
initially and $10$ batches of $q = 6$ observations 
are sequentially added. 
The objective functions are $50$ different sample 
realizations of a zero mean GP with unit variance and separable 
isotropic Mat\'ern(3/2) covariance function with range 
parameter equal to one. Both algorithms use the same 
initial design of experiment of $n$ points which 
are all S-optimal random Latin Hypercube designs 
\cite{kenny2000algorithmic}.
The mean and covariance function of the underlying GP 
are supposed to be known {(in practice, the hyperparameters of the GP model can be estimated by maximum likelihood \cite{D.Ginsbourger.etal2015})}. Since it is difficult to 
draw sample realizations of the GP on the whole 
input space $D := [0,1]^d$, we instead draw $50$ 
samples on a set of $2000$ space-filling 
locations and interpolate each sample in order 
to obtain the $50$ objective functions. 

Two variants of the BUCB algorithms are tested. 
Each of them constructs a batch by sequentially 
minimizing 
the kriging quantile $\mu^{\star}_{n}(\xx) - \beta_{n} s_{n}(\xx)$ where $s_{n}(\xx)=\sqrt{C_{n}(\xx,\xx)}$ is the posterior standard deviation at step $n$ and $\mu^{\star}_{n}(\xx)$ is the posterior mean conditioned both on the response at previous points and at points already selected in the current batch, with a dummy response fixed to their posterior means in the latter case. Following the settings of \cite{desautels2012parallelizing}, in the first and second variant of BUCB, the coefficients $\beta_{n}$ are given by:
\begin{align}
\beta_{n}^{(1)} := 2\beta_{\text{mult}} 
\log\left(\frac{\pi^2 d}{6 \delta}  (k+1)^2 \right) \text{ and }
\beta_{n}^{(2)} := 2\beta_{\text{mult}} 
\log\left(\frac{\pi^2 d}{6 \delta} (1+qk)^2 \right)
\end{align}
where $\beta_\text{mult} = 0.1$, $\delta=0.1$, and $k$ is 
the number of already evaluated batches at time $n$, i.e., here, $k\in\{0,\hdots,9\}$. The BUCB1 strategy is expected to select 
locations in regions with low posterior mean (exploitation) 
while BUCB2 is meant to favour more exploration due to 
a larger $\beta_n$. The minimization of the kriging quantile 
presented above is performed using a genetic algorithm 
\cite{genoud}.
Regarding the algorithm based on $\qEI$ sequential maximization, 
we propose to use a multi-start BFGS algorithm with 
analytical gradient. 
This algorithms operates gradient descents directly in the 
space of dimension $qd = 30$. 
To limit computation time, the number 
of starting batches in the multi-start is set to $3$. These 
$3$ batches are obtained by running the BUCB1 algorithm 
presented above with $3$ different values 
of $\beta_\text{mult}$ equals to $0.05, 0.1, 0.2$ 
respectively. 

At each iteration, we measure the regrets of each algorithm 
and average them over the $50$ experiments. 
To facilitate the interpretation of results, we first 
focus on the results of the algorithms after $1$ iteration, 
i.e. after having added only $1$ batch of $q$ points. 
We then discuss the results when $10$ iterations 
are run.
\subsubsection{First step of the optimization}
To start with, we focus on the selection of the first batch. 
 Table~\ref{table:iteration1} compares the average $\qEI$ and real improvement obtained for the three selection rules. For the first iteration only, the BUCB1 and BUCB2 selection rules are exactly the same. 
\begin{table}[h]
\centering
\begin{tabular}{m{3cm}>{\centering\arraybackslash}m{3cm}>{\centering\arraybackslash}m{3cm}}
Selection rule& Average expected improvement ($\qEI$)& Average realized improvement\\\hline
$\qEI$ & 0.672 & {\centering 0.697}\\
 BUCB   & 0.638& 0.638
\end{tabular}
\caption{Expected and observed first batch Improvement for $\qEI$ and BUCB batch selection methods, in average for 50 functions.}
\label{table:iteration1}
\end{table}
Since $\qEI$ is the one-step optimal, it is not a surprise that it performs better at iteration $1$ with our settings where the objective functions are sample realizations of a GP. If only one iteration is performed, improving the $\qEI$ 
is equivalent to improving the average performance. 
However, we point out that, in application, the maximization of $\qEI$ was not straightforward. It turns out that the batches proposed by the BUCB algorithms 
were excellent initial candidates in our descent algorithms.
The use of other rules for the starting batches, with points sampled uniformly 
or according to a density proportional to the one-point EI, did not manage to yield 
this level of performance.
\subsubsection{10 optimization steps}
The average regret of the different batch selection rules over $10$ iteration is depicted in Fig.~\ref{fig:regret}. 
This Figure illustrates that choosing the one-step optimal criterion is not necessarily optimal if more than one iteration is run \cite{ginsb_horizon}. 
Indeed, after two steps, $\qEI$ maximization is already beaten by BUCB2, and $\qEI$ becomes better again after iteration $7$.
Among the 50 optimized functions, $\qEI$ maximization gives the smallest 10-steps final regret for only 30\% of functions, against 52\% for the BUCB1 and 18\% for the BUCB2. 
On the other hand, the $\qEI$ selection rule is eventually better in average since,  for some functions, BUCB is beaten by $\qEI$ by a wide margin. This is further illustrated with the curve of the $95\%$ quantile of the regret which indicates that, for the worst 
simulations, $\qEI$ performs better. This gain in robustness alone explains the better average performance of $\qEI$.
Such improved performance comes at a price : 
the computational time 
of our multistart BFGS algorithm with  analytical gradient is $4.1$ times higher compared to the BUCB computation times.
\section{Conclusion}
In this article, we give a closed-form expression of the gradient of the multipoint Expected Improvement criterion, enabling an efficient $\qEI$ maximization at reduced computational cost.
Parallel optimization strategies based on maximization of $\qEI$ have been tested and are ready to be used on real test case with the DiceOptim R package. 
The BUCB algorithm turns out to be a good competitor to $\qEI$ maximization, with a lower computational cost, and also gives  
good starting batches for the proposed multistart BFGS algorithm. 
In general, however, the maximization of $\qEI$ remains a difficult problem. 
An interesting perspective is to develop algorithms taking advantage of some particular properties of the $\qEI$ function in the space of dimension $qd$, for example its invariance to point permutations. 
Other research perspectives include deriving cheap but trustworthy 
approximations of $\qEI$ and its gradient. 
Finally, as illustrated in the application, $\qEI$ sequential maximizations have no reason to constitute optimal decisions for a horizon beyond one batch. Although the optimal 
policy is known \cite{ginsb_horizon}, its implementation in practice remains an open problem.

\subsubsection*{Acknowledgement}
Part of this work has been conducted within the frame of the ReDice Consortium, gathering industrial (CEA, EDF, IFPEN, 
IRSN, Renault) and academic (\'Ecole des Mines de Saint-\'Etienne, INRIA, and the University of Bern) partners around advanced methods for Computer Experiments. The authors wish to thank Dylan Cable for reporting typos in the appendix.
\bibliographystyle{plain}
\bibliography{article}
\section{Appendix: Differential calculus}
\begin{itemize}
\item $\gan$ and $\gbn$ are functions giving respectively the mean of $\YY(\XX)$ and its covariance. Each component of these functions is either a linear or a quadratic combination of the trend function $\mmu$ or the covariance function $C$ evaluated at different points of $\XX$. The results are obtained by matrix differentiation. See the appendix B of \cite{ginsbourger:dicekrigingoptim} for a similar calculus.
\item $\gcn$ (resp. $\gdn$) is the affine (resp. linear) tranformation of the mean vector $\mm$ into $\mm^{(k)}$ (resp. the covariance matrix $\Sigma$ into $\Sigma^{(k)}$). The differentials are then expressed in terms of the same linear transformation :
\begin{equation}
\diff{\mm}{\gcn}(\hh)=L^{(k)}\hh\nonumber ~~\text{ and }~~\diff{\Sigma}{\gdn}(H)=L^{(k)}HL^{(k)\top}.\nonumber
\end{equation}
\item $\gen$ is defined by $\gen\left(\mm^{(k)},\Sigma^{(k)}\right)=\varphi_{\Sigma_{ii}^{(k)}}\left(m_i^{(k)}\right)$. Then the result is obtained by differentiating the univariate Gaussian probability density function with respect to its mean and variance parameters. Indeed we have :
\begin{align}
\diff{\left(\mm^{(k)},\Sigma^{(k)}\right)}{\gen}(h,H)=&~\diff{\mm^{(k)}}{\gen(\cdot,\Sigma^{(k)})}(h)+\diff{\Sigma^{(k)}}{\gen(\mm^{(k)},\cdot)}(H)\nonumber
\end{align}
\item $\gfn$ gives the mean and the covariance of $\ZZ^{(k)}_{-i}|Z_i=0$. We have :
\begin{equation}
\left( \mm^{(k)}_{|i},\Sigma^{(k)}_{|i} \right)=\gfn\left(\mm^{(k)},\Sigma^{(k)}\right)=\left( \mm^{(k)}_{-i}-\frac{m^{(k)}_i}{\Sigma_{ii}^{(k)}}\SSigma_{-i,i}^{(k)} ~, \Sigma_{-i,-i}^{(k)}-\frac{1}{\Sigma^{(k)}_{ii}}\SSigma_{-i,i}^{(k)}\SSigma_{-i,i}^{(k)\top}\right)\nonumber
\end{equation}
\begin{equation}
\diff{\left(\mm^{(k)},\Sigma^{(k)}\right)}{\gfn}(\hh,H) = \diff{\mm^{(k)}}{\gfn\left(\cdot,\Sigma^{(k)}\right)}(\hh)+\diff{\Sigma^{(k)}}{\gfn}\left(\mm^{(k)},\cdot\right)(H),\nonumber
\end{equation}
\begin{align}
&\text{with : }~~\diff{\mm^{(k)}}{\gfn\left(\cdot,\Sigma^{(k)}\right)}(\hh)= \left(\hh_{-i} -\frac{\hh_i}{\Sigma_{ii}^{(k)}}\SSigma_{-i,i}^{(k)} ~, ~0~\right)\nonumber\\
&\text{and : }~~\diff{\Sigma^{(k)}}{\gfn\left(\mm^{(k)},\cdot\right)}(H)=\left(\frac{m^{(k)}_i H_{ii}}{\Sigma_{ii}^{(k)2}}\SSigma_{-i,i}^{(k)}-\frac{m^{(k)}_i}{\Sigma_{ii}^{(k)}}H_{-i,i}~,\right.\nonumber\\ &\left.~H_{-i,-i}+\frac{H_{ii}}{\Sigma_{ii}^{(k)2}}\SSigma_{-i,i}^{(k)}\SSigma_{-i,i}^{(k)\top}-\frac{1}{\Sigma_{ii}^{(k)}}H_{-i,i}\SSigma_{-i,i}^{(k)\top}-\frac{1}{\Sigma_{ii}^{(k)}}\SSigma_{-i,i}^{(k)}H_{-i,i}\trans\right)\nonumber
\end{align}
\item $\ggn$ and  $\ghn$ : these two functions take a mean vector and a covariance matrix in argument and give a probability in output : $\Phi_{q,\Sigma^{(k)}}\left(-\mm^{(k)}\right)=\ggn\left(\mm^{(k)},\Sigma^{(k)}\right) $,
 $\Phi_{q-1,\Sigma^{(k)}_{|i}}\left(-\mm^{(k)}_{|i}\right)=\ghn\left(\mm^{(k)}_{|i},\Sigma^{(k)}_{|i}\right) $
So, for $\{p,\Gamma,\aa\} = \{q,\Sigma^{(k)},-\mm^{(k)}\}$ or $\{q-1,\Sigma^{(k)}_{|i},-\mm^{(k)}_{|i}\}$, we face the problem of differentiating a function $\Phi: (\aa,\Gamma)\to\Phi_{p,\Gamma}(\aa)$, with respect to $(\aa,\Gamma)\in \Rset^p\times\Scal_{++}^p$:
\begin{equation}
\diff{(\aa,\Gamma)}{\Phi}(\hh,H)=\diff{\aa}{\Phi(\cdot,\Gamma)}(\hh)+\diff{\Gamma}{\Phi(\aa,\cdot)}(H).\nonumber
\end{equation}
The the first differential of this sum can be written :
\begin{align}
\diff{\aa}{\Phi(\cdot,\Gamma)}(\hh) = \left\langle \left(\frac{\partial}{\partial a_i} \Phi(\aa,\Gamma)\right)_{1\le i\le p},\hh\right\rangle,\nonumber
\end{align}
with : $
\frac{\partial}{\partial a_i} \Phi(\aa,\Gamma) = \int \limits_{-\infty}^{a_1} \!\!\!\ldots\!\!\!\int \limits_{-\infty}^{a_{i-1}} \!\int \limits_{-\infty}^{a_{i+1}}\!\!\!\ldots\!\!\! \int \limits_{-\infty}^{a_p}  \varphi_{p,\Gamma}(u_{-i},a_i) \drm \uu_{-i}=\varphi_{1,\Gamma_{ii}}(a_i) \Phi_{p-1,\Gamma_{|i}}\left(\aa_{|i}\right). $
The last equality is obtained with the identity :
$\forall \uu \in \Rset^p,~ \varphi_{p,\Gamma}(\uu)=\varphi_{1,\Gamma_{ii}}(u_i) \varphi_{p-1,\Gamma_{|i}}(\uu_{|i}),$
with $\uu_{|i}=\uu_{-i}-\frac{u_i}{\Gamma_{ii}}\GGamma_{-i,i}$ and $\Gamma_{|i}=\Gamma_{-i,-i}-\frac{1}{\Gamma_{ii}}\GGamma_{-i,i}\GGamma_{-i,i}\trans$.
The second differential can be obtained via Plackett's formula \cite{berman1987extension}:
\begin{align}\diff{\Gamma}{\Phi(\aa,\cdot)}(H) := \tr\left(H . \left(\frac{\partial\Phi}{\partial \Gamma_{ij}} (\aa,\Gamma)\right)_{i,j\le p}\right)\nonumber
 = \frac{1}{2}\tr\left(H . \left(\frac{\partial^2\Phi}{\partial a_i\partial a_j}(\aa,\Gamma)\right)_{i,j\le p}\right).\nonumber
\end{align}
The second order derivatives can be calculated with the same approach as for the first order. We find: 

$
\frac{\partial^2\Phi}{\partial a_i\partial a_j}  (\aa,\Gamma)=
\left\lbrace
\begin{array}{lcc}
 \varphi_{2,\Gamma_{\{i,j\},\{i,j\}}}(a_i,a_j) \Phi_{p-2,\Gamma_{|ij}}(\aa_{|{ij}})\text{ , if }i\neq j,\nonumber\\
 -\frac{a_i}{\Gamma_{ii}} \frac{\partial}{\partial a_i}\Phi_{\Gamma}(\aa,\Gamma) - \sum_{\substack{j=1\\j\neq i}}^p \frac{\Gamma_{ij}}{\Gamma_{ii}}\frac{\partial^2}{\partial a_i\partial a_j}  \Phi(\aa,\Gamma)\nonumber \text{ otherwise}.
\end{array}\right.$
\end{itemize}
%
%
\clearpage
\end{document}